\documentclass[a4paper,USenglish,cleveref, autoref]{lipics-v2021}

\nolinenumbers
\pdfoutput=1 
\hideLIPIcs  

\usepackage{bm}
\usepackage[utf8]{inputenc}
\usepackage{xurl}
\usepackage{comment}
\usepackage{amssymb}
\usepackage{amsfonts}
\usepackage{graphicx}
\usepackage{url}
\usepackage{lstsemantic}
\usepackage{lstlangmizar}
\usepackage{booktabs}
\usepackage{footnoterange}
\usepackage{todonotes}
\usepackage{cite}
\usepackage{siunitx}
\usepackage{enigma}

\usepackage{hyperref}

\usepackage{doc}

\usepackage{paralist}

\title{The Isabelle ENIGMA}

\funding{
This work was partially supported by 
the
European Regional Development Fund under the Czech project AI\&Reasoning no. CZ.02.1.01/0.0/0.0/15\_003/0000466 (ZG, JP, JU), the ERC Starting
Grant \emph{SMART} no.~714034 (JJ, CK), Amazon Research Awards (JP, JU) and by the Czech MEYS under the ERC CZ project \emph{POSTMAN}
no.~LL1902 (JJ, JP).}
\author{Zarathustra A.~Goertzel}{Czech Technical University in Prague}{}{https://orcid.org/0000-0002-8458-2786}{}%

\author{Jan Jakub\r{u}v}{Czech Technical University in Prague and University of Innsbruck}{}{https://orcid.org/0000-0002-8848-5537}{}%

\author{Cezary Kaliszyk}{University of Innsbruck}{}{https://orcid.org/0000-0002-8273-6059}{}%

\author{Miroslav Olšák}{Institut des Hautes Études Scientifiques}{}{https://orcid.org/0000-0002-9361-1921}{}%
\author{Jelle Piepenbrock}{Czech Technical University in Prague and Radboud University}{}{https://orcid.org/0000-0002-8385-9157}{}%
\author{Josef Urban}{Czech Technical University in Prague}{}{https://orcid.org/0000-0002-1384-1613}{}%

\Copyright{Zarathustra A.~Goertzel, Jan Jakub\r{u}v, Cezary Kaliszyk, Miroslav Olšák, Jelle Piepenbrock and Josef Urban}


\authorrunning{Goertzel, Jakub\r{u}v, Kaliszyk, Olšák, Piepenbrock, Urban}

\titlerunning{The Isabelle ENIGMA}

\EventEditors{June Andronick and Leonardo de Moura}
\EventNoEds{2}
\EventLongTitle{13th International Conference on Interactive Theorem Proving (ITP 2022)}
\EventShortTitle{ITP 2022}
\EventAcronym{ITP}
\EventYear{2022}
\EventDate{August 7--10, 2022}
\EventLocation{Haifa, Israel}
\EventLogo{}
\SeriesVolume{237}
\ArticleNo{13}

\begin{document}

\maketitle

\begin{abstract}
  We significantly improve the performance of the E automated theorem
  prover on the Isabelle Sledgehammer problems by combining learning
  and theorem proving in several ways. In particular, we develop
  targeted versions of the ENIGMA guidance for the Isabelle problems,
  targeted versions of neural premise selection, and targeted
  strategies for E. The methods are trained in several iterations over
  hundreds of thousands untyped and typed first-order problems
  extracted from Isabelle. Our final best single-strategy ENIGMA and premise selection system
  improves the best previous version of E by 25.3\% %
  in 15 seconds,
  outperforming also all other previous ATP and SMT systems.
\end{abstract}

\section{Introduction}
\label{sect:introduction}

Formal verification in interactive theorem provers (ITPs) increasingly
benefits from general proof automation in the form of
\emph{hammers}~\cite{hammers4qed} and guided tactical
provers~\cite{GauthierKUKN21,BlaauwbroekUG20,NagashimaK17}.  In
particular, the Sledgehammer system~\cite{sledgehammer} for Isabelle
is today perhaps the most widely used strong general proof automation
system in ITP. %
In the recent years, machine learning and related AI methods for proof
automation have also been significantly
developed~\cite{AI4RFinalReport}. Such methods are relevant for
hammers in at least three ways: (i) learning-based \emph{premise
  selection}~\cite{abs-1108-3446,FarberK15,IrvingSAECU16,PiotrowskiU18,DBLP:conf/lpar/PiotrowskiU20,DBLP:conf/ecai/OlsakKU20}
usually improves the heuristic filters used by the hammers, (ii)
learning-based \emph{internal guidance} of the automated theorem
provers (ATPs) used for the heavy lifting in the hammers usually
improves on heuristic guidance of
ATPs~\cite{Veroff96,KaliszykU15,JakubuvU17a,KaliszykUMO18,JakubuvU19,DBLP:conf/cade/JakubuvCOP0U20,RawsonR21,GoertzelCJOU21},
and (iii) targeted theorem proving strategies developed by automated
strategy invention systems often improve on manually designed ATP
strategies~\cite{blistr,SchaferS15,JakubuvU17,HoldenK21}.

Most recent versions of such AI/TP methods have been developed
mainly on a %
fixed Mizar/MPTP
corpus~\cite{KaliszykU13b}, to allow easy comparisons with previously
developed methods. In particular, there the strongest 3-phase
single-strategy version of the ENIGMA system (based on E~\cite{Schulz13,Schulz19}) proves 56.35\% of the
holdout (test) toplevel theorems in 30s when using human-selected
premises~\cite{GoertzelCJOU21}. In higher time limits and by combining
human and learning-based premise selection, ENIGMA and Vampire~\cite{Vampire} today
prove 75\% of the toplevel Mizar
theorems.\footnote{\url{https://github.com/ai4reason/ATP_Proofs/blob/master/75percent_announce.md}}
These are good reasons for transferring the methods to other ITP
hammers.

A direct motivation for developing such AI/TP methods for Isabelle was
a recent request from the Sledgehammer developers for an
optimized version of ENIGMA for their GRUNGE-style~\cite{BrownGKSU19} evaluation of
multiple ATP systems and formats\cite{jasmin-seventeen}. While it was
not possible to do the work described in this paper on a two-week's
notice, it prompted us to start exporting and analyzing
the Isabelle datasets and developing suitable methods and systems for
them.

\subsection{Contributions}

We significantly improve the performance of the E automated theorem prover on the Isabelle Sledgehammer problems
by combining learning and theorem proving in several ways.
First, in Section~\ref{sect:isabelle} we extract two large datasets of
untyped first-order (FOF) and many-sorted first-order (TFF, TF0)
Isabelle Sledgehammer problems, using the Isabelle tool Mirabelle.
This results in almost 300000 aligned problems in each of the
exports, spanning in total 1902 Isabelle theory files and covering a
large number of topics in mathematics and formal verification. To our
knowledge, these are so far the largest corpora of Isabelle Sledgehammer
problems available today for training and evaluation of AI/TP systems.
Section~\ref{sect:incomplete} analyzes the corpora, showing that
they significantly differ from other large AI/TP datasets such as the
Mizar/MPTP toplevel theorems~\cite{KaliszykU13b} and the HOL4/GRUNGE
toplevel theorems\cite{BrownGKSU19}.

In Section~\ref{sect:strats}, we find optimized E strategies and
parameters for the corpora, which already improve on standard E on the
problems. They are suitable also for combinations with the ENIGMA
guidance, which is introduced in Section~\ref{sect:enigma}. We also
describe there several extensions to ENIGMA that were developed to
handle the Isabelle untyped and typed problems.  Section~\ref{sect:premsel} discusses
the neural premise selection that we use and its extensions for the typed
Isabelle setting.  Section~\ref{sect:eval} evaluates the methods in
several loops interleaving proving and learning from the proofs.
Our ultimate performance results are: (i) improving in 15s the original E
auto-schedule with the MePo filter by 25.3\%, %
when using a single
ENIGMA strategy with the best neural predictor, (ii) considerably
improving over all other ATPs and SMTs by a single ENIGMA strategy
combined with the best neural predictor, (iii) improving the
performance of all other systems by using the neural predictor, and
(iv) outperforming with ENIGMA all other ATPs and SMTs even when they are combined
with our %
predictor.

\section{Isabelle Problems}
\label{sect:isabelle}

To train and evaluate the Isabelle ENIGMA, we need
a dataset of Sledgehammer problems, which correspond to the proof obligations that users encounter
when using Isabelle as an interactive prover. We decided to focus on all
proof-intermediate goals visible to the users. This task has been tried as early
as in the first versions of the %
MPTP system \cite{Urb04-MPTP0}.
In Isabelle, it has been known as the ``Judgement Day''
evaluation, based on the paper with that title \cite{sledgehammer}.
We have used the Isabelle/Mirabelle infrastructure
to export all the problems encountered when building 179 Isabelle sessions. These
sessions originate from 75 sessions distributed with
Isabelle 2021-1, 80 selected sessions
from the AFP \cite{BlanchetteHMN15}, as well as 24 sessions distributed as part
of IsaFoR \cite{ceta}. All the sessions include in total 1902 Isabelle theory files.
The sessions with most problems can be categorized as
Analysis, Algebra,
Java Semantics, Category Theory,
Protocols, Term Rewriting, and Probability Theory with the largest 26 sessions listed in Table \ref{tab:largesessions}.

\begin{table}[htb!]
  \centering
  \begin{tabular}{ll|ll}
    \toprule
HOL-Nonstandard-Analysis & 1699 & Groebner-Macaulay & 4227 \\
Category2 & 1776 & HOL-ODE-Numerics & 4422 \\
Poincare-Bendixson & 1983 & HOL-MicroJava & 5183 \\
HOL-Number-Theory & 2071 & HOL-Auth & 5304 \\
MonoidalCategory & 2238 & HOL-Complex-Analysis & 5489 \\
HOL-Cardinals & 2268 & Groebner-Bases & 5710 \\
Core-DOM & 2280 & HOL-Computational-Algebra & 6280 \\
HOL-IMP & 2324 & Jordan-Normal-Form & 6786 \\
HOL-Data-Structures & 2353 & Category3 & 6818 \\
Dirichlet-Series & 2435 & HOL-Probability & 6954 \\
Slicing & 2517 & HOL-Decision-Procs & 7103 \\
HOLCF & 2524 & CR & 7341 \\
Formal-SSA & 2899 & HOL-Bali & 7804 \\
HOL-UNITY & 2938 & HOL & 7818 \\
HOL-Homology & 3022 & Goedel-HFSet-Semanticless & 8697 \\
HOL-ex & 3047 & HOL-Algebra & 9674 \\
CTRS & 3328 & HRB-Slicing & 10052 \\
HOL-Hoare-Parallel & 3733 & Jinja & 11520 \\
Signature-Groebner & 3762 & HOL-Library & 15627 \\
Valuation & 3786 & Bicategory & 16965 \\
Ordinary-Differential-Equations & 3885 & HOL-Nominal-Examples & 17145 \\
Smith-Normal-Form & 4045 & Group-Ring-Module & 19718 \\
Differential-Dynamic-Logic & 4158 & HOL-Analysis & 44172 \\ \bottomrule
  \end{tabular}
  \caption{The largest included sessions and their respective problem numbers}
  \label{tab:largesessions}
\end{table}

The Sledgehammer export allows multiple encodings of types, lambdas, and other options
\cite{BlanchetteBPS13}.
Since we are interested in the performance of learning-based first-order ATPs, we exported the problems in two first-order formats: TFF (also called TF0), i.e., many-sorted first-order logic, and FOF, i.e. untyped first-order logic.
For all problems we pre-selected 512 relevant premises using the heuristic MePo filter~\cite{MengP09}
before the translation.  This slightly overshoots the best performance
(256 premises) obtained by most of the top systems\footnote{Vampire is
  an exception: in~\cite{jasmin-seventeen} it is best with 512
  premises, likely due to its optimized SInE
  filter~\cite{HoderV11}.} on the FOF and TFF problems in the recent
Sledgehammer evaluation~\cite{jasmin-seventeen}. We use 512 premises because the
heuristic MePo filter is known to be weaker than state-of-the-art
selection systems (possibly pruning out some good premises too early),
and also because the 512-premise results of the best systems
in~\cite{jasmin-seventeen} are nearly identical\footnote{In particular, CVC5 - the winner in ~\cite{jasmin-seventeen} - is only 3.7\% (2626/2533) stronger with 256 premises.} to the 256-premise
results.\footnote{We could have used also 1024 premises, however already with 512 premises the datasets are becoming very large, making also the training of the ML systems technically challenging.}

For the other parameters, for E and Vampire we used the ones corresponding to the slice
selected when no-slicing is used for a particular prover. Additionally, when extracting
the FOF problems, we used the parameters used for such a slice in first-order E in the
previous Isabelle version. These parameters have been optimized by the Isabelle/Sledgehammer
developers based on experiments described in previous papers, e.g., \cite{BlanchetteBPS13}.
To ease comparison with \cite{jasmin-seventeen}, we use the polymorphic $g??$ \cite{jasmin-seventeen} encoding together with lambda-lifting~\cite{lambda-lifting} for FOF
and the native monomorphic encoding with lambda-lifting for TFF0.

Since the Mirabelle export has occasional problems with some
theories and encodings (theory compilation fails or does not
terminate with a particular export), we initially get different numbers of problems
for the FOF (293587) and TFF (386619) exports. To align the two
exports, we remove the non-overlapping problems, thus obtaining 276363
problems both in FOF and TFF that correspond to each other.  As usual
in machine learning, we then divide this dataset into the
\emph{training}, \emph{devel}opment (validation) and \emph{holdout}
(ultimate testing) parts. This is done by randomly shuffling the list
of the problems and dividing the shuffled list 90:5:5. This means that
we have 248727 problems to train our systems on, 13818 development
problems for controlling the hyperparameters of the learning and
building the best portfolios, and 13818 holdout problems on which the
trained systems are ultimately evaluated. We also sometimes use a 13818-big subset of the training set (\emph{small trains}). The total size of the FOF
dataset is about 50G compressed by gzip to 5.4G, while for the TFF
dataset it is about 90G, compressed by gzip to 7.7G. The complete datasets are publicly released at our accompanying
repository.\footnote{\url{https://github.com/ai4reason/isa_enigma_paper}}

The translation of the Isabelle/HOL problems to TPTP does not preserve
the names across the problems. The naming inconsistency can be as simple
as the naturals being given the constant name \texttt{nat} or
\texttt{nat2} in an encoded TPTP problem (this one happens because the
projection int-to-nat is also called \texttt{nat} in Isabelle),
depending on the order of defined constants in a given
problem. Additionally, Isabelle mangles names as part of the
encoding. For example in the basic theorem \texttt{List.distinct},
which states that an empty list is not equal to an applied list
constructor, an instance of the empty list can look like
\texttt{nil\_Pr1308055047at\_nat} for an empty list of products of
pairs of naturals.  This motivates our use of anonymous methods for
ENIGMA and premise selection in this work (Section \ref{sect:enigma},\ref{sect:premsel}).

\subsection{Differences to Related ITP/ATP Datasets}
\label{sect:incomplete}
The FOF and TFF Isabelle exports we use are intended to be sound but
generally sacrifice completeness to optimize ATP performance.
The possible sources of incompleteness include:
\begin{itemize}
\item The heuristic premise filter \cite{MengP09} pre-selecting only a fixed
  number of premises that are generally not guaranteed to justify
  the conjecture in Isabelle.
\item In the encodings, %
  polymorphic types (such as \texttt{'a list})
  are heuristically pre-instantiated
  (\emph{monomorphized}) by ground types. %
  This
  is an established optimization going back at least to Harrison's
  implementation of the MESON tactic~\cite{Har96} in HOL Light~\cite{Harrison96}, which
  can be seen as a particular kind of an abstraction step when
  reasoning in large theories.  Without a full abstraction-refinement
  loop~\cite{HernandezK18}, this is an obvious source of
  incompleteness, in a similar way as premise selection with a fixed premise limit.
\item Limited treatment of higher-order constructs such as lambda
  abstraction, typically not fully encoded in the FOF and TFF
  problems. The encodings %
  employ lambda-lifting, which is
  usually improving the ATP performance in practice, but is generally incomplete.
\end{itemize}

When developing new strategies, ATPs and premise selection
methods, such optimizations may be premature, having
different beneficial or adverse effects on the methods. In
particular, in the experiments conducted by us, we detect small
amount of incompleteness already with the baseline systems. For
example, CVC5 reports 256 problems in the whole TFF dataset to be
countersatisfiable. On the other hand, once a proof is found, it is typically
comparatively easy to replay from the minimized set of
premises by any ATP.

In this sense, the monomorphized Isabelle datasets considerably differ from other
datasets used for large AI/TP experiments such as the toplevel
theorems in the Mizar and HOL 4 libraries~\cite{BrownGKSU19}.
There, replaying the minimized proofs may still be quite hard for
ATPs, and the exports are typically striving for completeness, fully
delegating various abstraction-refinement methods such as
monomorphization and premise selection to the AI/TP systems that may
implement more complicated procedures for them.

We
measure this in more detail by comparing the clausified premise-minimized ATP problems solved by
Vampire and E on the Isabelle FOF dataset (88888 problems) and the
Mizar dataset (113332 problems) using several metrics computed in Table~\ref{MizVslsa}.  
\begin{table}[!htbp]
  \caption{\label{MizVslsa} Statistics of the Isabelle and Mizar
    clausified premise-minimized FOF problems solvable by E and
    Vampire. AC is the average number of clauses per problem, VC is
    the average number of clauses with variables per problem, EC is
    that for clauses with equality, iProver-10s is the number of
    problems solved by iProver limited to inst-gen calculus in 10s,
    and iProver-10s ratio is the ratio of that to the total number of
    problems.}
\begin{small}
\begin{center}
  \begin{tabular}{lllllll}
    \toprule
    Dataset & Problems & AC & VC &  EC &  iProver-10s & iProver-10s ratio\\
    \midrule                                    
    Isabelle FOF    &  88888 & 10.15 & 4.51  & 2.63 & 83015 & 0.93 \\
    Mizar             & 113332 & 35.55 & 23.16 & 10.31 & 65679  & 0.58        \\
    \midrule
    Ratio  Miz/Isa   &  & 3.50 & 5.14 & 3.92 &  &  0.62 \\
   \bottomrule
  \end{tabular}
\end{center}
\end{small}
\end{table}
The table shows that the number of clauses per
minimized problem is 3.5 times higher in Mizar. This may indicate the
difference between the (generally harder) toplevel ITP problems and the
intermediate goals. The most interesting difference is that about two
thirds of the clauses in the Mizar problems contain variables, while
in Isabelle this is only 44.4\% of the clauses.
Combined with the much
higher number of clauses in the Mizar problems, this leads to 5.14
times more clauses with variables in the Mizar problems. For clauses
with equality, this ratio is 3.92, i.e., also slightly higher than the
ratio of the clauses.
This means that the Isabelle problems are (after
minimization) much more ground and non-equational, and thus likely much more amenable to
instantiation-based methods than the Mizar problems. We confirm this
by running iProver \cite{DBLP:conf/cade/Korovin08} on both sets of minimized problems using only its Inst-Gen
calculus. In Mizar it solves 58\% of the problems while in Isabelle
93\%, i.e., 60\% more.

\section{Strategy Optimization for E and ENIGMA}
\label{sect:strats}
ATP
\emph{strategies} play a critical role when proving theorems. Their
targeted invention, optimization, and construction of their portfolios
(\emph{schedules}) may significantly improve the performance of the ATPs
in different domains. We have also found that some ATP
strategies behave better in combination with learning-based
guidance of the ATPs than others, and that it often seems
preferable to use a single strategy to produce the training data for
ENIGMA.\footnote{The use of single vs multiple strategies in
  combination with ENIGMA is not yet strongly experimentally
  explored. See, e.g.,~\cite{Goertzel20} for a recent related analysis.}

Our initial goals are thus to (i) find a strong set of E strategies
for the datasets, and in particular, to (ii) find a single strong E
strategy that behaves well in combination with the ENIGMA guidance.
We start exploring this on the FOF dataset, evaluating our 550
BliStr/Tune~\cite{blistr,JakubuvU17} strategies previously invented on
the Mizar, Sledgehammer, HOL, AIM and TPTP problems. This is done in two rounds.
In the first round, we run all the 553 strategies on a smaller sample of 500
randomly selected  FOF problems solvable by Vampire's CASC mode in 30
seconds.\footnote{We use here Vampire as a quick pre-filter for targeting the solvable problems by E strategies because in our preliminary experiments Vampire performed significantly better than E.}
After that, 
the 76 most performing and orthogonal strategies from the first run are evaluated on a bigger sample of 2000 Vampire-solvable problems.
This yields the following top 2 strategies in the greedy cover:
\begin{small}
\begin{verbatim}
protokoll_X----_auto_sine13 :995
protocol_eprover_f171197f65f27d1ba69648a20c844832c84a5dd7 :198
\end{verbatim}
\end{small}
The first strategy uses E's
auto-mode with a strong SInE filter, selecting up to 100 premises.
Unlike in the Mizar problems, the \texttt{hypos} parameter of SInE is
used here, giving the same importance to the local assumptions (TPTP role
\texttt{hypothesis}) as to the conjecture. We have
confirmed that this performs better than SInE without the
parameter on the problems. This leads us to construct
the ENIGMA features differently for Isabelle problems in Section~\ref{sect:enigma}.

The second %
strategy in the greedy cover (\texttt{f1711}) is the one
working best in the Mizar/MPTP setting, where it also
performs %
well when combined with the ENIGMA guidance. It is
however significantly weaker (921 vs 995 solved problems) than the
first auto-mode strategy. We conjecture that this is because it does not use SInE.
Adding a strong SInE filter (with the ``hypos''
parameter) indeed improves its performance to 1022 problems, making it
the strongest E strategy on the problems.  Since it %
is
also %
well behaved with the ENIGMA guidance, we use it in all
further experiments.
The base strategy (\texttt{f1711}) without any SInE filter will be denoted
as $\stratBase$, while the version with the SInE as $\stratSine$.
With the clausification changes explained next
we obtain two more strategies $\stratBaseDc$ and $\stratSineDc$.

\subsection{Clausification}

Clausification can have a large influence on the operation and
performance of ATPs.  In a setting with many complicated formulas,
naive clausification can lead to exponential blow-ups.
State-of-the-art ATPs counter that by introducing definitions for
subformulas. E's clausifier uses heuristic counting of the occurrences of each subformula to
decide when to introduce a new definition. The default factor (called \texttt{definitional-cnf}, \texttt{dc} for short)
for this used by E has been experimentally optimized to be 24 many years ago on the TPTP benchmark.
This may be however suboptimal for newer large-theory corpora, especially in encodings with type guards.
Also, a possible explanation for the relatively large improvement of E
by the aggressive SInE filter is that the clausification explodes
quite frequently on the unfiltered problems. We investigate this in several ways.

First, we simply try to clausify all FOF problems with the default E
options and a timeout of 60s.  This results in a gzipped total size of
21G, i.e. four times the size of the gzipped FOF problems.  This is
however without 28212 (10\% of all) problems that fail to get
clausified within 60s. This is a lot, because ITP hammers typically
give the ATPs a timeout of 15-30s to solve the whole problem.

This leads us to an experiment with smaller values for the definitional-cnf (\texttt{dc}) parameter on a sample of 1000 training problems. 
We use a 60s timeout for the clausification, measure the total size of gzipped cnfs, and the number of files where the clausification timed out. 
\begin{table}
  \caption{\label{CNFSize} Influence of the \texttt{dc} values on the clausification timeouts and size of the clausal problems.}
\begin{small}
\begin{center}
  \begin{tabular}{lllllll}
    \toprule
    definitional-cnf (\texttt{dc}) & 1 & 2 & 3 &  4 &  24 \\
    \midrule
    clausifications timed out in 60s (out of 1000) & 0 & 0 & 0 & 51 & 125  \\
    gzipped size of all clausified problems (MB)   & 36  & 47 & 163 & 120 & 77 \\
   \bottomrule
  \end{tabular}
\end{center}
\end{small}
\end{table}
The results are shown in Table~\ref{CNFSize}. The \texttt{dc} value of 3 is the
last one where there are no timeouts, but it already gives a 4-time
blowup over $\mathtt{dc}=2$.

Both more aggressive premise selection and more aggressive
introduction of new definitions can be used to counter the
clausification blowup on the Isabelle problems. Since the SInE filter
is only heuristic and usually inferior to trained premise selection,
we prefer more aggressive use of new definitions. To measure how much
the two methods interact, we evaluate our chosen 
strategy $\stratBase$ with and without SInE and with various \texttt{dc} values in
15s on our sample of 1000 problems.  The results are summarized in
Table~\ref{DC2Sine}. They confirm that the two methods interact a lot.
Setting $\mathtt{dc}=2$ replaces a lot of the improvement obtained by SInE with
the default $\mathtt{dc}=24$.  Since the SInE and non-SInE versions peak at
$\mathtt{dc}=3$ and $\mathtt{dc}=2$ respectively, we experiment with these values of
\texttt{dc} in our further experiments.
We denote $\stratBaseDc$ and $\stratSineDc$ the strategies obtained from
$\stratBase$ and $\stratSine$ by setting $\mathtt{dc}=3$.

\begin{table}
  \caption{\label{DC2Sine} 15s $\stratBase$ runs with/out SInE with different   \texttt{dc} values on 1000 sample problems.}
\begin{small}
\begin{center}
  \begin{tabular}{lllllll}
    \toprule
    definitional-cnf (\texttt{dc}) & 1 & 2 & 3 &  4 &  24 \\
    \midrule
    problems solved with SInE  & 242 & 268 & 271 & 266 & 263  \\
    problems solved without SInE    & 219  & 251 & 243 & 241 & 218 \\
   \bottomrule
  \end{tabular}
\end{center}
\end{small}
\end{table}

\section{ENIGMA for Isabelle }
\label{sect:enigma}

State-of-the-art automated theorem provers (ATP), such as E, Prover9, and
Vampire~\cite{Vampire}, are based on the saturation loop paradigm and the \emph{given clause
algorithm}~\cite{Overbeek:1974:NCA:321812.321814}.
The input problem, in first-order logic (FOF), is translated into a refutationally
equivalent set of clauses, and a search for contradiction is initiated.
The ATP maintains two sets of clauses: \emph{processed} (initially empty) and
\emph{unprocessed} (initially the input clauses).
At each iteration, one unprocessed clause is selected (\emph{given}), and all of the
possible inferences with all the processed clauses are generated
(typically using resolution, paramodulation, etc.), extending the unprocessed clause set.
The selected clause is then moved to the processed clause set.
Hence the invariant holds that all %
inferences among %
processed
clauses have been computed.

The selection of the ``right'' given clause is known to be vital for the
success of the proof search.
The first ENIGMA systems~\cite{JakubuvU17a,JakubuvU18,JakubuvU19,GoertzelJU19}
successfully implemented various ways of machine learning guidance for the
clause selection based on gradient boosting decision trees (GBDT).
Next generation 
ENIGMA~\cite{DBLP:conf/cade/ChvalovskyJ0U19,DBLP:conf/cade/JakubuvCOP0U20}
abstracts from symbol names with anonymization methods and additionally employs
graph neural network models (GNN) for clause selection.
The latest ENIGMA~\cite{GoertzelCJOU21} additionally implements clause
filtering of generated clauses (\emph{parental guidance}), and overcomes a
slower speed of GNN models with amortizing evaluation server.

\subsection{Model Training and Given Clause Guidance}

The training of ENIGMA models is usually done in a training/evaluation loop.
This general approach applies
both %
to clause guidance and
when filtering the generated clauses.

\begin{enumerate}
\item The training data $\sym{T}$ are gathered from a number of previous
   successful proof searches.  
   From each proof search, the training data consists of clauses processed
   during the proof search, labeled by flags \emph{positive} or \emph{negative}
   depending on whether they appear in the final proof.  
   These labeled clauses are translated to a suitable format for the underlying
   selection model (vectors for GBDT models, and tensors for GNNs).
\item Based on data $\sym{T}$, a GBDT (or a GNN model) $\sym{M}$ is trained.
   This model is capable of recognizing \emph{positive} clauses from 
   \emph{negatives} by assigning a score to an arbitrary clause.
\item The model $\sym{M}$ can be combined with an ordinary E's strategy
   $\sym{S}$ in a \emph{cooperative} way, yielding the ENIGMA strategy
   $\coop{\sym{S}}{\sym{M}}$.
   The ENIGMA strategy $\coop{\sym{S}}{\sym{M}}$ uses the model $\sym{M}$ to
   guide the given clause selection inside E, and it inherits other behaviour
   from $\sym{S}$.
   In the cooperative setting, about $50\%$ of the given clauses are selected
   as suggested by $\sym{M}$, while the remaining clauses are selected by the
   standard clause selection mechanism inherited from $\sym{S}$.
   Thus, ENIGMA compensates for a possible mistaken predictions of $\sym{M}$.
\item With new training data from new strategies, this process can be iterated.
\end{enumerate}

\subsection{Parental Guidance and Generated Clause Filtering}
\label{sect:pgintro}
ENIGMA models are applied within E in two capacities: 
(1) given clause selection and 
(2) parental guidance for filtering of the generated clauses. 
Clausal parental guidance evaluates a new clause $C$ based only on the features of the parents of $C$.
Parental guidance thus serves as a fast rejection filter: generated clauses
with scores below a chosen threshold are put into the \emph{freezer} set and
are only revived if E runs out of unprocessed clauses.  
Furthermore, such frozen clauses are never evaluated by other (possibly more
expensive) heuristics. This mechanism thus effectively (and in a complete way)
curbs the typically quadratic growth of the set of generated clauses.  
Full details can be found in previous work~\cite{GoertzelCJOU21} where it was
found that the the parental guidance is most effective when the concatenated
feature vectors of the parents are used as an input to the machine learning
model. 
The data for training parental guidance is generated by classifying parents of
proof clauses as \emph{positive} and all other generated clauses during a proof
search as \emph{negative}.  
To balance the data, the ratio of negative to positive examples is a valuable
hyperparameter.  

\subsection{Experiments with ENIGMA}

ENIGMA was so far used only with first-order logic (FOF) data in the TPTP
format.  In this work, we extend the usability of ENIGMA models also to simply
typed first-order formulas (TFF) of the TPTP format.
In the case of GBDTs models, we simply forget the type annotations.
Because GBDT ENIGMA models perform symbol name anonymization by replacing
symbol names by their arities, all the simple type names would get translated
to the same name anyway.
In the case of GNN models, we embed the type information in the clause graphs
by giving nodes representing variables of the same type by the same trained numerical representation (see
Section~\ref{sect:premsel}).

ENIGMA models embed information about the conjecture being proved inside
clause vectors/tensors. 
In this way, ENIGMA provides conjecture-specific suggestions.
The conjectures are marked in the input format with the TPTP role
\texttt{conjecture}.
In these experiments, we additionally treat clauses with the TPTP role
\texttt{hypothesis} just like conjectures.
This helps to further differentiate among various Isabelle problems.

In this work, we use ENIGMA GBDT models for clause guidance inside E (for given
clause selection and filtering of generated clauses), and we use the GNN models
only for the task of premise selection.
Section~\ref{sect:premsel} describes how the GNN models are used for
premise selection.
The experimental evaluation described in Section~\ref{sec:eval} presents the
results of training ENIGMA models for clause selection and parental guidance.

\section{Premise Selection for Isabelle via Graph Neural Networks}
\label{sect:premsel}
\label{sect:gnn}

A number of learning-based premise selection methods have been
developed for large ITP corpora and hammers in the last two decades. %
See~\cite{KuhlweinLTUH12-long,abs-1108-3446,hammers4qed,AI4RFinalReport}
for their overviews.  In a large evaluation done over the Mizar
corpus,\footnote{\url{https://github.com/ai4reason/ATP_Proofs}}
the strongest method turned out to be a property-invariant graph
neural network (GNN)
based on
the architecture previously used in several settings
\cite{DBLP:conf/ecai/OlsakKU20,DBLP:conf/cade/JakubuvCOP0U20,ZomboriUO21}.
We use this algorithm also for the Sledgehammer
problems here.

GNNs, and in particular this architecture preserve
several invariants of theorem proving data, such as insensitivity to
clause ordering and literal ordering. The inference (decisions) about which premises are relevant for a conjecture
are based on several rounds of neural message passing in a special graph constructed from the clauses corresponding to the formulas.
The property invariant
architecture also strives to be fully anonymous, in the sense that it
is invariant to all symbol names: the representations of symbols are
only based on their connectivity with other elements in the
formula. It also has a specific encoding for argument order that
allows the network to partially preserve this information and it has a
special handling of negation: terms of opposite polarity are related
by the corresponding operation $* -1$ in the float based
representation of the network.

This set of properties allows the
architecture to perform well in various theorem-proving settings. On
our Isabelle datasets, the symbol and name anonymity of the GNN is
particularly important.  As mentioned in Section~\ref{sect:isabelle},
the symbol names and the formula names are not used consistently here,
which would make the use of non-anonymous premise selection methods
difficult.
In this work, a 10-layer GNN was
used. The sizes of the first layer embeddings were 4, 1, 4 for the
\textit{term}, \textit{symbol} and \textit{clause} nodes
respectively. For the rest of the layers, the term, symbol and clause
nodes were represented by vectors of size 32, 64 and 32
respectively. The last, non-message passing layer that has the task of
predicting a probability for each premise had 128 neurons.

The GNN was newly modified to parse and make use of the typed TFF input. To take advantage of the
type information, we train separate embeddings for all types (2539 in Section~\ref{sect:GNN2TFF}) that
occur more than 10.000 times in the data. The GNN uses this type
embedding when reading in a variable, and the type embedding can
contain information about the type of the variable. Here, for simplicity, we chose to
directly learn the embeddings (initial GNN values before the start of the message passing) for the typed variable nodes.
This however does not fully preserve the
anonymity of the symbols in the GNN, which is one the core design principles of this neural
architecture. Adding instead an extra node in the GNN for each type would
allow us to preserve the anonymity also for types. In this setting the GNN would
learn to understand the types based only on their use in the current
problem, possibly thus generalizing better. This approach is however more complicated than our current solution and is left as future work here.

The Isabelle problems are big and their clausification by our GNN
parser may result in graphs with many clauses, even when
we heuristically pre-reduce the initial set of formulas proposed by
the MePo filter. This poses problems with the GPU memory (32 GB on our
machines) both during training the GNN and when using it for
predicting the relevant clauses. To counter that, we have introduced
several limits related to the number of nodes in the clause graphs
that allow us to skip very large clausified problems. The limit that
we currently use skips any problem that contains more than 50000 term
nodes after clausification (this corresponds roughly to the 95th percentile for the amount of term nodes in the problems).

\section{Evaluation}
\label{sec:eval}
\label{sect:eval}

We experiment with four variants of Isabelle problems.  
The first two are (1) $\expFOF$ and (2) $\expTFF$ without premise selection. 
Then there are two versions result from the GNN premise selector applied to the $\expTFF$
data: (3) $\expPREMA$ and (4) $\expPREMB$.

First, Section~\ref{sec:exp:lgb} describes experiments with %
given clause
guidance, and Section~\ref{sect:pgexp} describes experiments with adding parental guidance.
These two experiments were partially used to obtain the training data for
premise selection described in 
Section~\ref{sect:GNNTFF}
and 
Section~\ref{sect:GNN2TFF}.

\subsection{Evaluation of ENIGMA Given Clause Guidance}
\label{sec:exp:lgb}

We perform three separate evaluations of the GBDT (LightGBM~\cite{lightgbm}) ENIGMA clause
selection models on three different presentations of Isabelle problems.
(1) On the FOF translation (without premise selection) in Section~\ref{sec:exp:fof}, 
(2) on the TFF translation (without premise selection) in Section~\ref{sec:exp:tff},
and (3) on the TFF translation with GNN premise selection in Section~\ref{sec:exp:pre}.
The second premise selection dataset $\expPREMB$ is not used here. %

We experiment with combining training samples from different strategies.
Different E strategies might use different term orderings affecting the clause
normalization.
Since the ENIGMA %
models are syntax based, we only combine training
samples from \emph{compatible} strategies, which perform equivalent clause
normalization.
At this point, we consider strategies to be \emph{compatible} when they use the
same term ordering and literal selection function.

\subsubsection{Experiment $\expFOF$: First-Order Translation}
\label{sec:exp:fof}

\textbf{Setup.}
First, we experiment with the FOF translations of Isabelle problems without any
premise selection method applied.
E supports \emph{sine} filters to reduce the number of axioms of large
problems. 
Since the problems have no premise selection applied, 
we use two versions of the E strategy to obtain training problems: 
$\stratSine$ uses a manually selected sine filter\footnote{%
   \texttt{--sine='GSinE(CountFormulas,hypos,1.1,,03,20000,1.0)'}} 
and $\stratBase$ does not use a sine filter. 
We perform three training/evaluation loops as follows.
\begin{enumerate}
   \item \emph{Initial training data} $\sym{T}_0$: Evaluation of 
      $\stratBase$ and $\stratSine$ on the training problems.
   \item
      Train the model $\sym{L}$ on the current data $\sym{T}$.
   \item Evaluate $\coop{\stratBase}{\sym{L}}$ 
      and $\coop{\stratSine}{\sym{L}}$ on the training problems.
   \item Extend data $\sym{T}$ and continue with step $\mathbf{\mathsf{2}}$.
\end{enumerate}
We combine the two base strategies with model $\sym{L}$ in a cooperative way.
With model $\sym{L}$ we obtain two strategies with ENIGMA guidance, that is,
   $\coop{\stratBase}{\sym{L}}$
and
   $\coop{\stratSine}{\sym{L}}$.

\begin{table}[tb]
\begin{center}
\def\arraystretch{1.5}%
\setlength\tabcolsep{1.1mm}%
\begin{tabular}{l|l|l||r|r|r|r||r|r|r||r|r}
   &
   \multicolumn{2}{c||}{\textbf{notation}} & 
   \multicolumn{4}{c||}{\textbf{training}} & 
   \multicolumn{3}{c||}{\textbf{accuracy[\%]}} & 
   \multicolumn{2}{c}{\textbf{model}} \\
\hline
   \textit{l}  & 
   \textit{trains}  & 
   \textit{model}  &
   \textit{probs}  & 
   \textit{proofs}  & 
   \textit{rows} &
   \textit{filesize}  &
   \textit{acc}  & 
   \textit{pos}  & 
   \textit{neg}  &
   \textit{time}  & 
   \textit{filesize} \\
\hline

0 & $\symFOF{T}_{0}$ & $\symFOF{L}_{0}$ & 
70K & 114K &
8M &
1.1G &
92.8 &
89.8 &
93.4 &
0:12 &
54.8M \\

1 & $\symFOF{T}_{1}$ & $\symFOF{L}_{1}$ & 
81K & 255K &
16M &
2.3G &
87.8 &
82.1 &
89.0 &
0:20 &
54.9M \\

2 & $\symFOF{T}_{2}$ & $\symFOF{L}_{2}$ & 
84K & 400K & 
23M &
3.2G&
85.6 &
81.9 &
86.5 &
0:31 &
55.1M \\

\end{tabular}
\end{center}
\caption{Experiment \expFOF: Learning statistics (Section~\ref{sec:exp:fof}).}
\label{tab:fof:ml}
\end{table}

\textbf{Learning Statistics.} 
Table~\ref{tab:fof:ml} presents training data statistics and models evaluation
for the three training/evaluation loops performed in this $\expFOF$
experiments.
There is:
\begin{itemize}
   \item \textbf{training}: The column \emph{probs} is the number of training
      problems in the training data, while the column \emph{proofs} is the
      number of different successful proof runs, where we can have multiple
      proofs for a single problem.  The column \emph{rows} signifies the number
      of vectors in the training data, each vector corresponding to one clause
      in the proofs.  The column \emph{filesize} is the file size of the
      \emph{compressed} training samples.
   \item \textbf{accuracy}: Columns \emph{acc}, \emph{pos}, \emph{neg} show
      testing accuracies of each model on the testing set in percents.  Column
      \emph{acc} show the overall model accuracy, while columns \emph{pos} and
      \emph{neg} show testing accuracy on positive and negative testing samples
      separately.
   \item \textbf{model}: The column \emph{time} shows the time needed for model
      training (in hours and minutes), and the column \emph{size} shows the
      LightGBM model file size.  Model file size is an important suggestion of
      the model ATP performance, since the model size influences the model
      loading time and prediction times in E.
\end{itemize}

When training a model, we set aside $5\%$ of the training data 
in order to compute the testing \textbf{accuracy}.  
The model is trained on the remaining $95\%$.
\footnote{
   The numbers in the \textbf{training} columns are only on the training $95\%$ subset.} 
This split is done on the level of solved problem names rather than on proofs
or vectors so that all the proofs of a single problem will appear either in the $95\%$
training subset, or all in the $5\%$ testing subset.
This is important to keep the testing set unbiased.
Otherwise, the testing data can partially overlap with the training data, since
two proofs of the same problem tend to be quite similar.
This split on solved problem names is computed independently in every loop iteration.
This split is done only on the training problems of the global
training/development/holdout split used for further experiment in this paper.

From the numbers in Table~\ref{tab:fof:ml}, we can see that the number of
solved problems (column \emph{probs}) in the data increases with every loop
iteration but much more slowly than the value in the column \emph{proofs}.
This means that we are obtaining duplicate proofs for already solved problems,
since we include all the proofs for all solved problems in the training data in
this experiment.
Note that the testing accuracies decrease with increasing training data size.
All the models have been built in less than $30$ minutes and result in a
similarly sized model file.
Also note that number of \emph{proofs} grows much faster than the problems
solved (\emph{probs}).
It shows that we often prove the same problems.

\begin{table}[tb]
\begin{center}
\def\arraystretch{1.2}%
\setlength\tabcolsep{1.5mm}%
\begin{tabular}{l|ll||r|r|r|r||r|r|r|r}
   &
   \multicolumn{2}{c||}{\textbf{strategy}} & 
   \multicolumn{4}{c||}{\textbf{trains solved by}} & 
   \multicolumn{4}{c}{\textbf{devels solved by}} \\
\hline
   \textit{l}  & 
   \textit{base}  & 
   \textit{sine}  &
   \textit{base}  & 
   \textit{sine}  & 
   \textit{both} &
   \textit{total}  &
   \textit{base}  & 
   \textit{sine}  & 
   \textit{both} &
   \textit{total}  \\
\hline

- & $\sym{S}_\n{base}$ & $\sym{S}_\n{size}$ & 
   \num[math-rm=\mathbf]{56921} & 
   \num[math-rm=\mathbf]{65124} &
   \num[math-rm=\mathit]{75080} &
   \num[math-rm=\mathit]{75080} &
   \num[math-rm=\mathbf]{3114} &
   \num[math-rm=\mathbf]{3567} &
   \num[math-rm=\mathit]{4084} &
   \num[math-rm=\mathit]{4084} \\

0 & \multicolumn{2}{c||}{$\coop{\sym{S}_\star}{\symFOF{L}_0}$} & 
   \num[math-rm=\mathbf]{77084} & 
   \num[math-rm=\mathbf]{72869} &
   \num[math-rm=\mathit]{85903} & 
   \num[math-rm=\mathit]{86661} &
   \num[math-rm=\mathbf]{3888} &
   \num[math-rm=\mathbf]{3886} &
   \num[math-rm=\mathit]{4552} &
   \num[math-rm=\mathit]{4784} \\

1 & \multicolumn{2}{c||}{$\coop{\sym{S}_\star}{\symFOF{L}_1}$} & 
   \num[math-rm=\mathbf]{80613} &
   \num[math-rm=\mathbf]{74191} &
   \num[math-rm=\mathit]{87734} & 
   \num[math-rm=\mathit]{89886} &
   \num[math-rm=\mathbf]{3933} &
   \num[math-rm=\mathbf]{3851} &
   \num[math-rm=\mathit]{4516} &
   \num[math-rm=\mathit]{4947} \\

2 & \multicolumn{2}{c||}{$\coop{\sym{S}_\star}{\symFOF{L}_2}$} & 
   \num[math-rm=\mathbf]{81640} &
   \num[math-rm=\mathbf]{74878} &
   \num[math-rm=\mathit]{88566} & 
   \num[math-rm=\mathit]{91261} &
   \num[math-rm=\mathbf]{3963} &
   \num[math-rm=\mathbf]{3894} &
   \num[math-rm=\mathit]{4558} &
   \num[math-rm=\mathit]{5036} \\

\end{tabular}
\end{center}
\caption{Experiment \expFOF: ATP performance (Section~\ref{sec:exp:fof}).}
\label{tab:fof:atp}
\end{table}

\textbf{ATP Evaluation.} Table~\ref{tab:fof:atp} shows the ENIGMA models
performance separately on training (\textbf{trains}) and on development
problems (\textbf{devel}).
Since the development problems were not used during the training in any way, this
evaluation tells how much are the ENIGMA model over-fitting on the training
files.

Every row describes the performance of two strategies specified in the column
\textbf{strategy}.
Problems solved by the two strategies individually are in the first two
\textbf{bold} columns.
\textit{Italics} values display a total cover of set of strategies.
The column \emph{both} 
shows the number of problems solved both by the two strategies together.
This is helpful to estimate the complementarity of \emph{base} and \emph{sine}
strategies.
Two strategies are \emph{complementary}, when they solve different
problems.
The column \emph{total} shows the cumulative number of problems solved 
by all the current strategies (above in the table).

In Table~\ref{tab:fof:atp}, we see that the \emph{sine} strategy performed 
better than \emph{base} initially.
However, from the first learning the \emph{base} strategy dominates.
This suggests that ENIGMA learns to do premise selection on its own to some
extent (when trained on the samples from the \emph{sine} strategy).
All \emph{base} and \emph{sine} strategies are, however, quite complementary.
In total, we start with \num{75080} solved problems and we end up with \num{91261}
after the learning, almost $22\%$ improvement on trains
($23\%$ on devels).
The best single strategy is improved by $25\%$ on trains
(and by $11\%$ on devels).

It is interesting to observe, how the \emph{base} strategies in one iteration
improves on both \emph{base} and \emph{sine} from the previous iteration, as if
merging the two strategies into one.
It suggests that additional proof samples from compatible
but complementary strategies
could lead to an additional improvement.
We further investigate this %
in the next experiment (Section~\ref{sec:exp:tff}).

\subsubsection{Experiment $\expTFF$: Typed First-Order Formulae}
\label{sec:exp:tff}

\textbf{Setup.}
We perform a similar experiment as for the FOF, but this time targeted to the
TFF Isabelle translation.
\begin{enumerate}
\item
Again, we start with the training data obtained by the evaluation of
$\stratBase$ and $\stratSine$.
\item We run three iterations of the
training/evaluation loop.
\item After the three iterations, we additionally evaluate two
more pure E strategies $\stratBaseDc$ and $\stratSineDc$ which improve on
$\stratBase$ and $\stratSine$ by adjusting E's clausification algorithm 
(switch E's option ``\texttt{definitional-cnf}'' from $24$ to $3$).
\item We perform two more training/evaluation loops with the expanded training data.
\end{enumerate}

\begin{table}[tb]
\begin{center}
\def\arraystretch{1.5}%
\setlength\tabcolsep{1.2mm}%
\begin{tabular}{l|c|c||r|r|r|r||r|r|r||r|r}
   &
   \multicolumn{2}{c||}{\textbf{notation}} & 
   \multicolumn{4}{c||}{\textbf{training}} & 
   \multicolumn{3}{c||}{\textbf{accuracy[\%]}} & 
   \multicolumn{2}{c}{\textbf{model}} \\
\hline
   \textit{l}  & 
   \textit{trains}  & 
   \textit{model}  &
   \textit{probs}  & 
   \textit{proofs}  & 
   \textit{rows} &
   \textit{size}  &
   \textit{acc}  & 
   \textit{pos}  & 
   \textit{neg}  &
   \textit{time}  & 
   \textit{size} \\
\hline

0 & $\symTFF{T}_{0}$ & $\symTFF{L}_{0}$ & 
108K & 186K &
10,3M &
1.2G &
89.6 &
86.2 &
90.2 &
12:36 &
54.8M \\

1 & $\symTFF{T}_{1}$ & $\symTFF{L}_{1}$ & 
114K & 383K &
19,6M &
2.2G &
85.0 &
78.8 &
86.2 &
20:29 &
55.0M \\

2 & $\symTFF{T}_{2}$ & $\symTFF{L}_{2}$ & 
117K & 587K & 
27,9M &
3.1G &
82.6 &
77.4 &
83.8 &
20:52 &
55.1M \\

\hline
\hline
3 & $\symTFF{T}_{3}$ & $\symTFF{L}_{3}$ & 
122K & 822K &
39,3M &
4.3G &
81.4 &
77.8 &
82.2 &
23:17 &
55.2M \\

4 & $\symTFF{T}_{4}$ & $\symTFF{L}_{4}$ & 
123K & 1.03M &
48,6M &
5.3G &
80.9 &
77.6 &
81.7 &
29:46 &
55.3M \\

\end{tabular}
\end{center}
\caption{Experiment $\expTFF$: Learning statistics (Section~\ref{sec:exp:tff}).}
\label{tab:tff:ml}
\end{table}

\textbf{Learning Statistics.}
Table~\ref{tab:tff:ml} presents the machine learning evaluation 
(in the same format as Table~\ref{tab:fof:ml} described in Section~\ref{sec:exp:fof}).
Before the fourth loop ($l=3$), we additionally evaluate all the strategies
$\coop{\sym{S}}{\sym{L}}$,
for $\sym{S}$ ranging over $\stratBaseDc$ and $\stratSineDc$, and 
for $\sym{L}$ ranging over the models of the first three loops. 
This gives us additional training data for the fourth iteration, reflected in
the table by a sudden increase in both solved problems (\emph{probs}) and
\emph{proof} count (in the row $l=3$).
We see similar training times and model sizes as in the FOF experiment.

\begin{table}[tb]
\begin{center}
\def\arraystretch{1.2}%
\setlength\tabcolsep{1.5mm}%
\begin{tabular}{l|ll||r|r|r|r||r|r|r|r}
   &
   \multicolumn{2}{c||}{\textbf{strategy}} & 
   \multicolumn{4}{c||}{\textbf{trains solved by}} & 
   \multicolumn{4}{c}{\textbf{devels solved by}} \\
\hline
   \textit{l}  & 
   \textit{base}  & 
   \textit{sine}  &
   \textit{base}  & 
   \textit{sine}  & 
   \textit{both} &
   \textit{total}  &
   \textit{base}  & 
   \textit{sine}  & 
   \textit{both} &
   \textit{total}  \\
\hline

- & $\sym{S}_\n{base}$ & $\sym{S}_\n{size}$ & 
   \num[math-rm=\mathbf]{100259} &
   \num[math-rm=\mathbf]{98317} & 
   \num[math-rm=\mathit]{114838} &
   \num[math-rm=\mathit]{114838} &
   \num[math-rm=\mathbf]{5532} &
   \num[math-rm=\mathbf]{5403} &
   \num[math-rm=\mathit]{6347} &
   \num[math-rm=\mathit]{6347} \\

0 & \multicolumn{2}{c||}{$\coop{\sym{S}_\star}{\symTFF{L}_0}$} & 
   \num[math-rm=\mathbf]{108377} &
   \num[math-rm=\mathbf]{101271} &
   \num[math-rm=\mathit]{118262} &
   \num[math-rm=\mathit]{121353} &
   \num[math-rm=\mathbf]{5468} &
   \num[math-rm=\mathbf]{5347} &
   \num[math-rm=\mathit]{6222} &
   \num[math-rm=\mathit]{6692} \\

1 & \multicolumn{2}{c||}{$\coop{\sym{S}_\star}{\symTFF{L}_1}$} & 

\num[math-rm=\mathbf]{113729} &
\num[math-rm=\mathbf]{103382} &
\num[math-rm=\mathit]{121995} &
\num[math-rm=\mathit]{124795} &
   \num[math-rm=\mathbf]{5788} &
   \num[math-rm=\mathbf]{5471} &
   \num[math-rm=\mathit]{6466} &
   \num[math-rm=\mathit]{6894} \\

2 & \multicolumn{2}{c||}{$\coop{\sym{S}_\star}{\symTFF{L}_2}$} & 

\num[math-rm=\mathbf]{115790} &
\num[math-rm=\mathbf]{104270} & 
\num[math-rm=\mathit]{123400} &
\num[math-rm=\mathit]{126344} &
   \num[math-rm=\mathbf]{5934} &
   \num[math-rm=\mathbf]{5505} &
   \num[math-rm=\mathit]{6547} &
   \num[math-rm=\mathit]{6894} \\

\hline

* & $\sym{S}_\n{base3}$ & $\sym{S}_\n{sine3}$ & 
\num[math-rm=\mathbf]{106132} &
\num[math-rm=\mathbf]{100904} &
\num[math-rm=\mathit]{118925} &
\num[math-rm=\mathit]{132552} &
   \num[math-rm=\mathbf]{5881} &
   \num[math-rm=\mathbf]{5522} &
   \num[math-rm=\mathit]{6515} &
   \num[math-rm=\mathit]{7160} \\

\hline

3 & \multicolumn{2}{c||}{$\coop{\sym{S}_{\star3}}{\symTFF{L}_3}$} & 

\num[math-rm=\mathbf]{122492} & 
\num[math-rm=\mathbf]{107035} &
\num[math-rm=\mathit]{127955} &
\num[math-rm=\mathit]{133222} &
   \num[math-rm=\mathbf]{6293} &
   \num[math-rm=\mathbf]{5673} &
   \num[math-rm=\mathit]{6785} &
   \num[math-rm=\mathit]{7280} \\

4 & \multicolumn{2}{c||}{$\coop{\sym{S}_{\star3}}{\symTFF{L}_4}$} & 

\num[math-rm=\mathbf]{122931} & 
\num[math-rm=\mathbf]{107316} & 
\num[math-rm=\mathit]{128339} &
\num[math-rm=\mathit]{133762} &

   \num[math-rm=\mathbf]{6277} &
   \num[math-rm=\mathbf]{5704} &
   \num[math-rm=\mathit]{6812} &
   \num[math-rm=\mathit]{7326} \\

\end{tabular}
\end{center}
\caption{Experiment \expTFF: ATP performance (Section~\ref{sec:exp:tff}).}
\label{tab:tff:atp}
\end{table}

\textbf{ATP Evaluation.}
Table~\ref{tab:tff:atp} presents the ATP evaluation 
(in the same format as Table~\ref{tab:fof:atp} described in Section~\ref{sec:exp:fof}).
As opposed to the FOF experiment, the \emph{base} strategies dominate from the
beginning.
Both strategies are still highly complementary.
The evaluation of $\stratBaseDc$ and $\stratSineDc$ strategies boosts the
number of solved trains from \num{126344} to \num{132552}.
This highly improves the performance of the best strategy (\emph{base}) in the
fourth iteration ($l=3$) from \num{115790} to \num{122492}, that is, by $5.8\%$.
It shows that additional external training data can be quite useful during the
training.
We further investigate this issue in the next experiment (Section~\ref{sec:exp:pre}).

\subsubsection{Experiment $\expPREMA$: First GNN Premise Selection}
\label{sec:exp:pre}

\textbf{Setup.}
Here we experiment with GNN premise selection data $\expPREMA$ obtained by
applying GNN premise selection to the $\expTFF$ problems.
The GNN premise selection produces several collections of the training problems
(called \emph{slices}) with a slightly different clause selection criterion.
We experiment with two slices $\sliceMinus$ and $\sliceSix$, which were
experimentally found well performing and complementary.
Our first experiment is aimed at generating a large collection of training
samples.
\begin{enumerate}
   \item We perform three loops of training/evaluation, just as in 
      the TFF experiment, separately on $\sliceMinus$ and $\sliceSix$.
      We loop with the base strategies $\stratBaseDc$ and $\stratSineDc$.
   \item 
      We merge the training data from the previous two separate experiments
      and perform three more loops on the merged data.
      However, we drop the \emph{sine} strategies and evaluate only the strategy
      $\coop{\stratBaseDc}{\sym{L}}$ on the two $\expPREMA$ slices.
   \item
      From the above we collect a large database of \num{108}K proved training
      problems.
      Since the collection can contain duplicate proofs of a single problem, we
      select just three proofs per problem.
      We use the proof pos/neg ratios as a measure of proof similarity, and
      select proofs thusly different.
   \item 
      The training data from the last step, denoted $\symPREMA{T}_{\n{three}}$,
      gives us one final model $\symPREMA{L}_{\n{three}}$.
\end{enumerate}
Next experiment tries to gather even more training samples.
\begin{enumerate}
   \item We additionally consider training data from the previous TFF experiments.
\item We gather even more valuable training samples from ENIGMA parental
   guidance experiments on slices $\sliceMinus$ and $\sliceSix$.
\item We select three proofs per problem from $\expTFF$ samples.
\item We select three proofs per problem from $\expPREMA$ samples.
\item The training data $\symPREMA{T}_{\n{sixes}}$ contain six
   proofs per problem and yield model $\symPREMA{L}_{\n{sixes}}$.
\end{enumerate}

\begin{table}[tb]
\begin{center}
\def\arraystretch{1.5}%
\setlength\tabcolsep{1.2mm}%
\begin{tabular}{c|c||r|r|r|r||r|r|r||r|r}
   \multicolumn{2}{c||}{\textbf{notation}} & 
   \multicolumn{4}{c||}{\textbf{training}} & 
   \multicolumn{3}{c||}{\textbf{accuracy[\%]}} & 
   \multicolumn{2}{c}{\textbf{model}} \\
\hline
   \textit{trains}  & 
   \textit{model}  &
   \textit{probs}  & 
   \textit{proofs}  & 
   \textit{rows} &
   \textit{size}  &
   \textit{acc}  & 
   \textit{pos}  & 
   \textit{neg}  &
   \textit{time}  & 
   \textit{size} \\
\hline

$\symPREMA{T}_{\n{three}}$ & $\symPREMA{L}_{\n{three}}$ & 
108K & 186K &
10,3M &
1.2G &
89.6 &
86.2 &
90.2 &
12:36 &
54.8M \\

$\symPREMA{T}_{\n{six}}$ & $\symPREMA{L}_{\n{six}}$ & 
133K & 763K &

28,6M &
3.26G &
77.6 &
76.8 &
78.0 &
25:44 &
77.9M \\

\end{tabular}
\end{center}
\caption{Experiment $\expPREMA$: Learning statistics (Section~\ref{sec:exp:pre}).}
\label{tab:pre:ml}
\end{table}

\textbf{Learning Statistics.}
Table~\ref{tab:pre:ml} presents the machine learning evaluation (in the same
format as Table~\ref{tab:fof:ml} described in Section~\ref{sec:exp:fof}).
Note the huge difference in the number of \emph{proofs}, resulting in 
much larger \emph{training data size}.
The second training data include proofs of more than \num{25}K additional
problems (\emph{probs}).
Training times and model sizes clearly reflect the training file size. 

\begin{table}[tb]
\begin{center}
\def\arraystretch{1.5}%
\setlength\tabcolsep{1.5mm}%
\begin{tabular}{l||r|r|r|r||r|r|r|r}
   & 
   \multicolumn{4}{c||}{\textbf{trains solved by}} & 
   \multicolumn{4}{c}{\textbf{devels solved by}} \\
\hline
   \textit{strategy}  & 
   $\sliceMinus$  & 
   $\sliceSix$  & 
   \textit{both} &
   \textit{total}  &
   $\sliceMinus$  & 
   $\sliceSix$  & 
   \textit{both} &
   \textit{total}  \\
\hline

$\sym{S}=\sym{S}_\n{base3}$ & 
   \num[math-rm=\mathbf]{122196} &
   \num[math-rm=\mathbf]{117341} &
   \num[math-rm=\mathit]{126323} &
   \num[math-rm=\mathit]{126323} &

   \num[math-rm=\mathbf]{6706} &
   \num[math-rm=\mathbf]{6462} &
   \num[math-rm=\mathit]{6955} &
   \num[math-rm=\mathit]{6955} \\

$\coop{\sym{S}}{\symPREMA{L}_{\n{three}}}$ &
   \num[math-rm=\mathbf]{127606} &
   \num[math-rm=\mathbf]{120248} &
   \num[math-rm=\mathit]{129971} &
   \num[math-rm=\mathit]{132431} &

   \num[math-rm=\mathbf]{6800} &
   \num[math-rm=\mathbf]{6495} &
   \num[math-rm=\mathit]{6994} &
   \num[math-rm=\mathit]{7251} \\

$\coop{\sym{S}}{\symPREMA{L}_{\n{six}}}$ &

   \num[math-rm=\mathbf]{132063} &
   \num[math-rm=\mathbf]{123229} &
   \num[math-rm=\mathit]{134544} &
   \num[math-rm=\mathit]{135823} &
   \num[math-rm=\mathbf]{6994} &
   \num[math-rm=\mathbf]{6591} &
   \num[math-rm=\mathit]{7153} &
   \num[math-rm=\mathit]{7380} \\

\end{tabular}
\end{center}
\caption{Experiment $\expPREMA$: ATP performance (Section~\ref{sec:exp:pre}).}
\label{tab:pre:atp}
\end{table}

\textbf{ATP Evaluation.}
Table~\ref{tab:pre:atp} presents the ATP evaluation 
(in the same format as Table~\ref{tab:fof:atp} described in Section~\ref{sec:exp:fof}).
Here, however, we evaluate the single strategy $\coop{\stratBaseDc}{\sym{L}}$
on slices $\sliceMinus$ and $\sliceSix$, instead of using two \emph{base} and
\emph{sine} strategies.

Firstly, we note the effect of the premise selection itself.
The performance on $\stratSineDc$ improved by more than $15\%$ from the
previous experiment (from \num{106132} to \num{122196}).
The model $\symPREMA{L}_{\n{three}}$ performs quite well, being trained on
proofs \num{108}K problems, it solves almost \num{128}K problems.
The model $\symPREMA{L}_{\n{six}}$ further boosts the performance, showing that
combining of training data from various compatible sources might be beneficial.
Comparing the performance on trains with the performance on devels, we can
conclude that ENIGMA LightGBM clause selection models slightly overfit
but they are still capable of %
generalization.

\subsection{Evaluation of the Parental Guidance}
\label{sect:pgexp}
\textbf{Setup.} Parental guidance models are co-trained with clause selection models in a series of loops over the training data.  
In each loop iteration, the LightGBM %
parameters for parental guidance models are tuned using a series of grid searches with Optuna~\cite{optuna_2019}. 
These are the number of leaves, the bagging fraction and frequency, the minimum number of samples to create a new leaf, and L1 and L2 regularization.  
The learning rate is fixed at \num{0.15}, the maximum tree depth is capped at \num{256} and the number of trees is \num{250}.  
The number of leaves is varied between \num{256} and \num{3333}. %
The best result of each grid search is used for the next parameter's grid search. 
Accuracy on positive training examples is considered twice as important as the accuracy on negatives when choosing which parameters perform best. 
There are multiple reasons for this.  A primary reason is that the confidence in positive examples is higher than confidence for the classification of negatives 
because a negative clause in one successful proof search could be positive in another proof search. 
The resulting model is evaluated with the nine parental filtering thresholds, $\{0.03, 0.05, 0.1, 0.2, 0.3, 0.4, 0.5, 0.6, 0.7\}$, 
over a set of \num{300} problems from the development set for \num{30} seconds.  
This is done for the vanilla TFF problems as well as the problems with premise selection slices. %
The best run (as evaluated by a greedy cover) on each version of the problems is then run on the full training set.  
Then the problems from runs in the total greedy cover are used as data for the next iteration of looping.  
This means that some problems can have over \num{10} proofs in the training data.

\textbf{Iterations.}
The training of parental guidance was done with the aim to develop as strong a performance as possible, using diverse data. 
The models for loops $\sym{L}_{1}$ and $\sym{L}_{2}$ are run and trained on the TFF data that do not use
premise selection.  Models $\sym{L}_{3}$ 
and $\sym{L}_{4}$ are only run on the small trains set. %
The models $\sym{L}_{3}$ to $\sym{L}_{8}$ are run on $\expPREMA$.  
Finally, models $\sym{L}_{9}$ and $\sym{L}_{10}$ are run on $\expPREMB^{-1}$.  
The largest performance jumps correspond to the addition of premise selection  (Table~\ref{PGTable}). 
The strongest parental guidance models are always on the $\textrm{PRE}^{-1}$ premise selection data 
and the $\expPREMA^{64}$ slices provide fewer complementary problems than the baseline TFF problems.

\textbf{The best model} $\sym{L}_{10}$, with the second premise selection slices, $\expPREMB^{-1}$, proves \num{168} problems ($56\%$) in \num{30}s on the parameter tuning development set, and 
\num{137893} problems ($55.4\%$) on the training set in \num{15}s.  
In \num{30}s, $\sym{L}_{10}$ proves \num{7472} problems ($54.1\%$) on the development set and \num{7466} problems ($54\%$) on the holdout.  
Without premise selection, $\sym{L}_{10}$ proves \num{133390} training problems ($53.6\%$), 
which indicates that training on the premise selection data transfers back to the original problems. %
The remaining results are presented in Table~\ref{pg:res}.  
This parental+ENIGMA model is our final product. 
It solves \num{7395} holdout problems in \num{15}s, thus significantly improving over unguided E and also over all other ATPs and SMTs. 
It also outperforms all other ATPs and SMTs even when they use our best premises (Table~\ref{GNNEval1}).

\begin{table}[tb]
\begin{center}
\def\arraystretch{1.5}%
\setlength\tabcolsep{1.5mm}%
\begin{tabular}{l|r|r||r|r|r|r|r|r||r|r}
   &
   $\sym{L}_{1}$ &
   $\sym{L}_{2}$ &
   $\sym{L}_{3}$ &
   $\sym{L}_{4}$ &
   $\sym{L}_{5}$ &
   $\sym{L}_{6}$ &
   $\sym{L}_{7}$ &
   $\sym{L}_{8}$ &
   $\sym{L}_{9}$ &
   $\sym{L}_{10}$ 
   \\ \hline
small trains & 6475 & 6718 & 7081 & 7140 & 7312 & 7351 & 7407 & 7417 & 7647 & \bf{7705}
   \\ \hline
devel & 6241 & 6462 & 6928 & 6892 & 6566 & 6850 & 7070 & 7115 & 7277 & \bf{7379}
   \\ \hline
holdout & 6251 & 6459 & 6886 & 6843 & 6581 & 6816 & 7015 & 7062 & 7352 & \bf{7395}
\end{tabular}
\end{center}
\caption{\label{PGTable}Parental guidance iterations on small trains, devel, and holdout (\num{13818} problems in \num{15}s). 
         Loops $\sym{L}_{1}$ and $\sym{L}_{2}$ are run on TFF data, $\sym{L}_{3}$ to $\sym{L}_{8}$ on $\expPREMA^{-1}$,
         and $\sym{L}_{9}$ and $\sym{L}_{10}$ on $\expPREMB^{-1}$.}
\label{pg:res}
\end{table}

\subsection{First Training of Premise Selection on TFF Problems ($\expPREMA$)}
\label{sect:GNNTFF}

We have done several large experiments with the GNN-based premise
selection (Section~\ref{sect:premsel}), first in the untyped and then
in the typed setting.  For lack of space we include below only the two final
experiments on the TFF data, where most of the ENIGMA runs were done.

For the first round of training the GNN on the TFF data we are using
the proof data produced only by the base sine/nosine TFF runs of unguided
E and the first three ENIGMA iterations on the TFF training set.
Altogether these runs produce 1701284 proof dependencies.
These dependencies are first deduplicated to 353875, and then we also
for every problem
$P$ remove all premise sets subsumed by a smaller premise set.
This further decreases the size of the dataset to 242432 proof dependencies, for 131309 unique solved problems.
Most of the solved problems (80993) have after this redundancy elimination only one solution, while for the remaining ones we get from 2 to 16 different solutions. Since problems with 1 to 3 solutions dominate the dataset (163650 out of the total 242432 solutions), we do not do further pruning of over-represented proofs for the training (as in the next training run).

For this first training and prediction we do not yet use the new typed
extensions of the GNN. Instead, all TFF formulas are stripped of their
type information and given to the network as untyped FOF.
Each problem uses its original conjecture, the positives are the
premises used in a given proof and the negatives are all other
premises for that problem (i.e., all the MePo premises). This sometimes leads to large training
inputs, so we normalize them to have size at most 500KB by randomly
removing negatives. The whole training dataset has  size 46GB.
We then train the GNN on it with batch size 10,  learning rate 0.005, and with balancing the loss on the positive and negative premises.

The training for two full epochs on an NVIDIA Volta 100 takes about 12
hours, saving the weights 16 times.  The balanced accuracy increases
from 0.8533 to the final 0.9067 (0.9061 vs 0.9073 on positives vs
negatives) in our final snapshot, which we then use for prediction
over all 276363 problems. This is parallelized over four GPUs, taking
several hours. For each problem we use the predictions to produce 5
premise selections based on the GNN score threshold (1,0,-1,-2,-3,-4),
and 5 premise selections based on old-style top slices of the ranked
premises (16,32,64,128,256). We do a small search with 200 development
problems and the base strategy over this grid, which is won by the
-1-based predictions, best complemented by the 64-based
predictions.
These premise selections are denoted $\sliceMinus$ and $\sliceSix$ in the other parts of this paper.
E/ENIGMA are then evaluated on both of them, while we
also evaluate other systems only on the -1-based predictions (Table~\ref{GNNEval1}).

\subsection{Second Training of Premise Selection on TFF Problems ($\expPREMB$)}
\label{sect:GNN2TFF}

The second premise selection training is done by the typed version of
the GNN (Section~\ref{sect:premsel}), using explicitly 2539 types that occur with frequency higher
than 10000 in the training data. The remaining types (over 300000 in
the training set) %
are mapped to
the same generic embedding, which means that the GNN treats them all as the same type.  The overhead for the 2539 distinguished
most frequent types increases the size of the GNN only by 100kb.
The training uses again a batch size of 20 and a learning rate of 0.0005.
The training dataset is created from all TFF training problems solved in the previous loops, both by E/ENIGMA and CVC5 and Vampire.
This gives 823141 unique premise selections for 146576 solved problems.
The 823141 unique premise selections are again minimized with
respect to subsumption,
reducing them to 488186 minimal premise selections.  To address the imbalance
caused by having various numbers of proofs for a single problem in the
training set, we keep at most three proofs for each problem. This
further reduces the set to 292080 examples. The examples are
again all reduced to a size of at most 500KB.

This results in our final training set with an overall size of about 60GB.
Since the reduction of the TFF inputs does not generally guarantee to prevent a blow-up during the clausification,
we also further use here a size limit of 50000 nodes inside the GNN parser (Section~\ref{sect:premsel})
and filter out such large graphs which may otherwise deplete the GPU memory.
The GNN is trained for full two epochs on the data, taking about one
day on a single NVIDIA V100 GPU and storing the weight files 15 times per epoch.
For producing the final predictions, we take the 28th weights with the
highest balanced accuracy of 0.9221 (0.9391 / 0.9051 for
positives/negatives). We produce the same grid of predictions as in
the first round for all problems. The -1-based predictions are again
the winner, best complemented by the 0-based predictions.
These premise selections are denoted $\sliceMinusB$ and $\sliceZeroB$ in the other parts of this paper.
Table~\ref{GNNEval1} shows that also all non-ENIGMA systems benefit from the GNN predictions, and that the second round
improves over the first round of predictions for all of
them.

\begin{table}
  \caption{\label{GNNEval1} Final comparison with non-ENIGMA systems: E2.6 with its auto-schedule, CVC5, and Vampire-CASC (master 4909). Each run standalone (MePo predictions) and with the first/second -1 GNN TFF predictions. The last entry is the final/best $Loop_{10}$ (parental) ENIGMA (Section~\ref{sect:pgexp}).}
\begin{small}
\begin{center}
  \begin{tabular}{lllllll}
    \toprule
    method & E auto-sched. & CVC5 & Vampire                      & $\sym{L}_{10}$ ENIGMA\\
    \midrule
    15s devel, no premsel & 5891 & 7053 & 6452                   & 7133 \\
    15s holdout, no premsel   & 5903  & 7051 & 6454              & 7139 \\
     30s holdout, no premsel   & 6089  & 7140 & 6945             & 7170 \\
        15s devel, preds -1 (1st round) & 6968 & 7211 & 7023     & 7191   \\
    15s holdout, preds -1  (1st round)  & 6956  & 7158 & 6978    & 7155 \\
        15s devel, preds -1 (2nd round) & 7074 & 7394 & 7132     & 7379 \\
    15s holdout, preds -1  (2nd round)  & 7066  & 7372 & 7118    & \bf{7395} \\
     30s holdout, preds -1  (2nd round)  & 7139  & 7398 &  7397  & \bf{7466} \\
    \bottomrule
  \end{tabular}
\end{center}
\end{small}
\end{table}

\section{Conclusion}

We have developed versions of the ENIGMA systems and neural premise selectors for
the Isabelle Sledgehammer problems.  Our best single-strategy system using the parental ENIGMA guidance and the typed GNN premise selection
solves 7395 holdout problems in 15s, 
improving on original E's auto-schedule performance (5903) by 25.3\%. %
It also improves on all other ATPs and SMTs, both when used standalone and when
used in conjunction with our best neural premise selection.
To achieve this, we have produced large corpora of Isabelle problems
for training and evaluation of the AI/TP methods, and developed new extensions of our
systems, especially for the typed setting. 

\newpage
\bibliographystyle{plain}
\bibliography{ate11}

\appendix

\end{document}